\documentclass{ieeeaccess}
\usepackage{cite}
\usepackage{amsmath,amssymb,amsfonts}
\usepackage{graphicx}
\usepackage{textcomp}

\usepackage{booktabs}
\usepackage{algorithm}
\usepackage[noend]{algpseudocode}

\usepackage{bm}
\makeatletter
\AtBeginDocument{\DeclareMathVersion{bold}
\SetSymbolFont{operators}{bold}{T1}{times}{b}{n}
\SetSymbolFont{NewLetters}{bold}{T1}{times}{b}{it}
\SetMathAlphabet{\mathrm}{bold}{T1}{times}{b}{n}
\SetMathAlphabet{\mathit}{bold}{T1}{times}{b}{it}
\SetMathAlphabet{\mathbf}{bold}{T1}{times}{b}{n}
\SetMathAlphabet{\mathtt}{bold}{OT1}{pcr}{b}{n}
\SetSymbolFont{symbols}{bold}{OMS}{cmsy}{b}{n}
\renewcommand\boldmath{\@nomath\boldmath\mathversion{bold}}}
\makeatother

\def\BibTeX{{\rm B\kern-.05em{\sc i\kern-.025em b}\kern-.08em
    T\kern-.1667em\lower.7ex\hbox{E}\kern-.125emX}}

\begin{document}
\doi{10.1109/ACCESS.2024.0429000}

\title{ViSIR: Spectral Bias Mitigation via Vision Transformer–Tuned Sinusoidal Implicit Networks for ESM Super-Resolution}
\author{\uppercase{ehsan zeraatkar}\authorrefmark{1},
\uppercase{Salah A. faroughi}\authorrefmark{2}, and \uppercase{Jelena Te\v{s}i\'c}\authorrefmark{3}}

\address[1,3]{Computer Science, Texas State University, San Marcos, 78666, Texas}
\address[2]{Chemical Engineering, University of Utah, Salt Lake City, 84112, Utah}



\corresp{Corresponding author: Ehsan Zeraatkar (e-mail: ehsanzeraatkar@txstate.edu)}

\begin{abstract}\\
\textbf{Purpose:} Earth system models (ESMs) integrate the interactions of the atmosphere, ocean, land, ice, and biosphere to estimate the state of regional and global climate under a wide variety of conditions. The ESMs are highly complex; thus, deep neural network architectures are used to model the complexity and store the down-sampled data. This paper proposes the Vision Transformer Sinusoidal Representation Networks (ViSIR) to improve the ESM data's single image SR (SR) reconstruction task. 

\textbf{Methods:} ViSIR combines the SR capability of Vision Transformers (ViT) with the high-frequency detail preservation of the Sinusoidal Representation Network (SIREN) to address the spectral bias observed in SR tasks. 

\textbf{Results:} The ViSIR outperforms SRCNN by 2.16 db, ViT by 6.29 dB, SIREN by 8.34 dB, and SR-Generative Adversarial (SRGANs) by 7.93 dB PSNR on average for three different measurements.

\textbf{Conclusion:} The proposed ViSIR is evaluated and compared with state-of-the-art methods. The results show that the proposed algorithm is outperforming other methods in terms of Mean Square Error(MSE), Peak-Signal-to-Noise-Ratio(PSNR), and Structural Similarity Index Measure(SSIM).
\end{abstract}


\titlepgskip=-21pt

\maketitle

\section{Introduction}
\label{sec1}
\PARstart{A}{n} Earth System Model (ESM) is a highly advanced computer model that consolidates the interactions among the various components of the Earth—e.g., the atmosphere, ocean, land, ice, and biosphere—to provide minute information about the climate and state of the environment of the Earth. By providing accurate descriptions of physical, chemical, and biological cycles, ESMs enable researchers to study multidimensional processes of climate change \cite{climate2013ems}. These models represent phenomena ranging from atmospheric circulation and ocean currents to land surface processes, ice movement, and vegetation patterns and capture a holistic description of the Earth system.

ESMs are advanced compared to simple climate models as they incorporate additional processes, such as the carbon cycle, which is liable for the strongest feedback on the physical climate. The increased complexity makes ESMs necessary in projecting climate change's impact, such as sea level increase, weather disasters, and water resources \cite{heinze2019}. But Earth System Models (ESMs) attempts the simulation of coupled dynamics of atmosphere, oceans, land surface and cryosphere over the whole world, solving an enormous number of partial differential equations for fluid motion, radiation, chemistry and biogeochemical cycles on a three-dimensional grid. To keep the computational expenses feasible, even on supercomputers available today, such grids have horizontal resolutions in the range of 50–200 km and a few dozen vertical layers only. Consequently, such processes at scales less than those mentioned above-by way of example, convective storms, urban heat-island effects, topographic influences at small scales, and localized surface fluxes-have to be "parameterized" (i.e., approximated by bulk formulas) instead of being treated explicitly. Scaling up such crude, highly global model outputs to the scale of a county or municipality can be difficult because typical decision makers require resolution better than 1 km and the original model grid cells must be interpolated or aggregated, smearing gradients in temperature, precipitation, soil moisture and pollutant concentrations. This blurring not only reduces the realism of local extremes such as flash floods or heatwaves but also introduces very significant uncertainty for stakeholders who lack the particular expertise or computing power to run their own high-resolution regional or convection-permitting models. \cite{eyring2016, vandal2017}.

Super Resolution (SR) is thus an effective solution to this problem. In the context of computer vision, SR is a name given to techniques that enhance image resolution by transforming low-resolution (LR) images into high-resolution (HR) images \cite{yang2019}. SR has widespread applications, from surveillance and medical imaging to media content enhancement \cite{Dong2016}. Traditional SR techniques have, nonetheless, failed to effectively capture fine, high-frequency details \cite{ledig2017, tai2017}.

The traditional approaches are ineffective in the SR task \cite{ledig2017,tai2017}. The super-resolution problem we address in this paper is transforming a high-resolution ESM image using the corresponding low-resolution image and a model. The model we propose in this paper is a new hybrid model and an extension of the Vision Transformer (ViT) and Sinusoidal Representation Network (SIREN). Although ViT was a good tool that could capture long-range relationships in images\cite{touvron2021, Dosovitskiy2020}, it may not be enough to capture high-frequency components essential to get high-quality Super Resolution\cite{bai2022}. The SIRENs have demonstrated that they can capture high-frequency details in images\cite{SIREN}, and here we present ViSIR. This mixed network replaces the final fully connected layer of ViT with SIREN to address the spectral bias issue in the SR tasks of the Earth System Model images.

\subsection{Main Contributions}

\textbf{Hybrid ViT–SIREN Architecture for ESM SR.} We introduce ViSIR, the first method to embed a Sinusoidal Representation Network (SIREN) directly into the final layers of a Vision Transformer (ViT), thereby combining ViT’s global‐context modeling with SIREN’s high‐frequency detail recovery.

\textbf{Frequency‐Tuned Implicit Representation.} We propose a novel hyperparameter search over SIREN’s frequency parameter~$\omega_0$ within a transformer pipeline—mitigating spectral bias more effectively than standalone SIREN or ViT.

\textbf{Ablation Study of Model Components.} We systematically isolate the contributions of (i) the ViT backbone, (ii) the SIREN module, and (iii) our novel integration design, confirming that their combination drives the performance improvements reported.

\section{Related Work}
\label{sec-related}
\PARstart{T}{he} advent of Convolutional Neural Networks (CNNs) revolutionized super-resolution (SR) reconstruction, as evidenced in early research such as SRCNN \cite{Dong2016}. CNNs learn low-to-high resolution image mappings by effectively extracting slowly varying and smooth features. Still, they fail to recover fine details and abrupt intensity changes, giving rise to the issue of spectral bias \cite{Zhang2019}. In SRCNN, color layers and channels are optimized simultaneously to create improved quality results compared to the conventional interpolation methods. Then, intensive comparisons among SRCNN, Fast SRCNN-ESM, Efficient Sub-pixel CNN, Enhanced Deep Residual Network, and SRGANs \cite{Nikhil2024} revealed that deeper residual structures like Enhanced Deep SR (EDSR) \cite{Kim2016EDSR} are better in providing PSNR and better in restoring high-frequency parts of Earth System Model (ESM) images.

Progress in the depth of networks has led to methods such as Very Deep SR (VDSR) \cite{Kim2016VDSR} and Residual Dense Networks (RDN) \cite{Zhang2018}. These networks exploit residual learning and dense connections to extract hierarchical features from LR images but sometimes suffer from over-smoothing by pixel-wise loss functions such as Mean Squared Error (MSE). To improve reconstruction quality further, researchers have turned towards implicit neural representations. The generalized INR (GINR) network \cite{grattarola2022ginr} approximates discrete sample points using spectral graph embeddings, and the Higher-Order INR (HOIN) approach \cite{chen2024hoin} employs neural tangent kernels to induce high-order interactions among features and mitigate spectral bias.

Deep generative models have also played a significant role in SR tasks. Techniques based on SRGAN \cite{shidqi2023} have been useful in downsampling climate data and transforming LR ESM data into HR images for regional precipitation predictions. Multimodal methods integrating numerical weather prediction models and attention to U-Net have also been utilized to improve temperature predictions \cite{Ding2024}.

The Sinusoidal Representation Network (SIREN) offers an additional complementary method, utilizing periodic activation functions to recover lost high-frequency details in normal CNN processing \cite{SIREN}. The frequency bias-resolving capability of SIREN also directly helps the output of SR, especially for images with a complex texture.

In more contemporary times, transformer models have been strong competitors in the SR domain. The Vision Transformer (ViT) \cite{Dosovitskiy2020} splits images into patches and uses multi-head self-attention for capturing long-range relations and global context, with improved efficiency compared to the traditional CNNs \cite{Irani2025}. However, despite their promise, transformers are extremely data- and computationally intensive. Researchers have created optimized transformer architectures specifically for SR applications to overcome this.

For instance, Zhong et al. \cite{Zhong2023} investigated transformer models for spatial downscaling and bias correction in weather forecasts and demonstrated that models like SwinIR and Uformer outperform traditional CNNs in spatial detail preservation. Building on this, Karwowska and Wierzbicki \cite{Karwowska2024} introduced a novel ESRGAN with Uformer blocks to enhance video satellite imagery with remarkable improvement in global and local quality metrics.

Ma et al. \cite{Ma2025} proposed DESRGAN, a detail-enhanced generative adversarial network for small sample SISR in the GAN area. DESRGAN restores texture and edge details and relieves overfitting using a shallower generator model. Furthermore, Guo et al. \cite{Guo2025} introduced a learnable adaptive bilateral filter to achieve better generalization in SISR to offset the synthetic-real distribution gap of LR images. Wu et al. \cite{Wu2025} proposed an unsupervised dual contrastive learning-based super-resolution model that leverages unpaired data and cycle consistency for enhanced latent feature learning and reconstruction accuracy to complement supervised approaches.

Collectively, these advancements—from CNN-based methods to transformer-enhanced and GAN-based techniques—demonstrate the rapid evolution of SR methodologies. They emphasize the importance of overcoming spectral bias, preserving high-frequency details, and achieving robust generalization across diverse real-world conditions. Our work builds on these advances by integrating CNNs, transformers, and innovative loss functions to address the unique challenges posed by ESM data, offering improved performance for both single-image and video super-resolution tasks.

Section~\ref{sec-method} introduces the methodology behind \textbf{ViSIR}, the hybrid vision transformer algorithm. In Section~\ref{sec-data}, we describe the ESM dataset and how the RGB image collection was derived, and the Proof of Concept in Section~\ref{sec-POC} for the three experiments and the summary of findings we present in Section~\ref{sec-Exp}.

\begin{figure}[!ht]
 \centering
 \includegraphics[width=\linewidth]{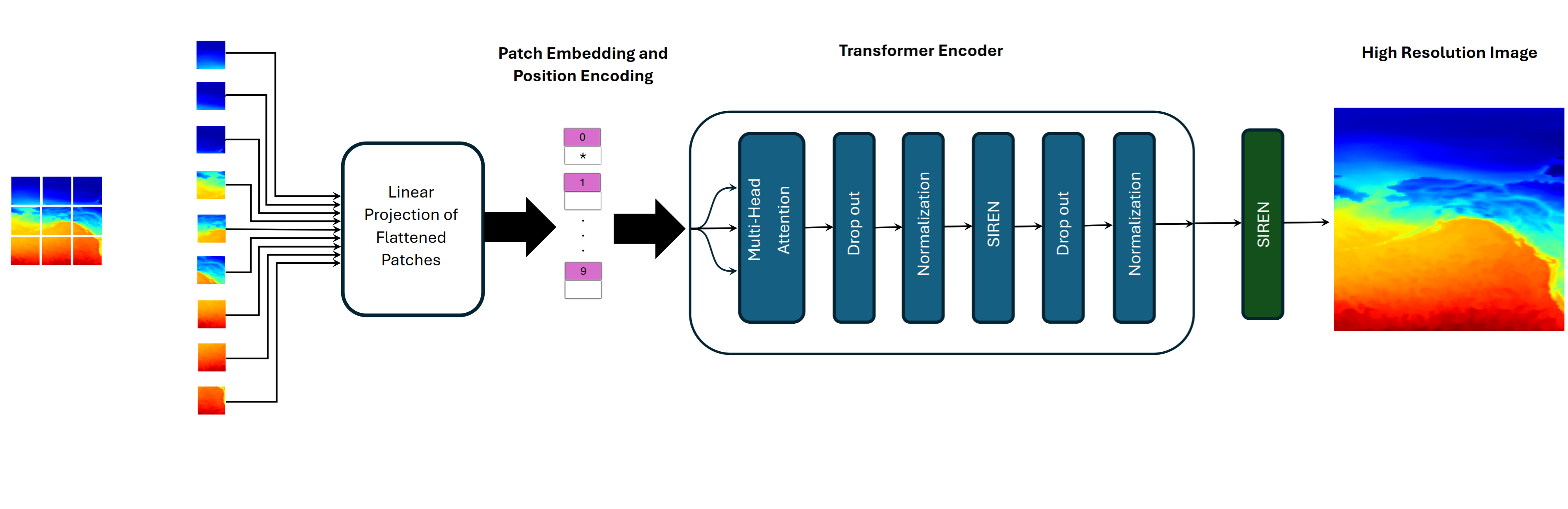} 
 \vspace*{-2em}
 \caption{ViSIR divides the input image into patches, pre-processes them using embedding and position encoding, and feeds the input to a visual transformer followed by the SIREN architecture.}
 \label{fig-Flowchart}
 \vspace*{-1em}
\end{figure}

\section{Methodology}
\label{sec-method}
\PARstart{V}{iSIR} pipeline flow is illustrated in Figure~\ref{fig-Flowchart}. The ViT processes the input images in the proposed pipeline to capture global dependencies and essential contextual information for Super Resolution. Given a low-resolution image \( I_{\text{LR}} \), the patch embedding operation can be described as:

\begin{equation}
E_i = W_p \cdot \text{patch}(I_{\text{LR}})_i + b_p,\quad i=1,\dots,N,
\label{eq-patch_embedding}
\end{equation}

\noindent where \( W_p \) and \( b_p \) are the patch embedding weights and bias, and \( N \) is the number of patches. These embeddings are then processed through multiple transformer layers, yielding contextualized features:

\begin{equation}
T = \text{ViT}(I_{\text{LR}}).
\label{eq-transformer_output}
\end{equation}

\noindent The SIREN then uses this information to address the spectral bias in the input image, resulting in higher-resolution output, as outlined in Eq.~\ref{eq-siren_layer}

\begin{equation}
f(x) = \sin\left(\omega_0 W x + b\right),
\label{eq-siren_layer}
\end{equation}

\noindent The \( \omega_0 \) is a hyperparameter controlling the frequency of the sinusoidal function. In ViSIR, the final reconstruction is performed by feeding the transformer’s output into one or more SIREN layers, as outlined in the Eq~\ref{eq-final_reconstruction}, 
where \( W_r \) and \( b_r \) are the weights and bias of the SIREN module. 

\begin{equation}
R = f(T) = \sin\left(\omega_0 W_r T + b_r\right),
\label{eq-final_reconstruction}
\end{equation}

Next, ViSIR integrates ViT's global context modeling capabilities with SIREN's high-frequency representation strength. Algorithm 1 presents a concise pseudo-code overview of the proposed ViSIR methodology, detailing the steps from patch extraction and token embedding through transformer-based processing to the final reconstruction using a SIREN layer.

\begin{algorithm}
  \caption{ViSIR Super‑Resolution}
  \begin{algorithmic}[1]
    \State \textbf{Input:} Low-resolution image $LR$, patch size $P$, number of layers $L$, number of heads $H$, embedding dimension $D$, frequency $\omega_0$
\State \textbf{Output:} Reconstructed high-resolution image $HR$
\State Divide $LR$ into non-overlapping patches of size $P \times P$.
\For{each patch}
    \State Compute a linear embedding to obtain token $x \in \mathbb{R}^D$.
\EndFor
\State Form token sequence $X = \{x_1, x_2, \dots, x_N\}$ and add positional encodings.
\For{$\ell = 1$ to $L$}
    \State Apply multi-head self-attention on $X$ using $H$ heads.
    \State Add a residual connection and perform layer normalization.
    \State Process the result through a SIREN network.
\EndFor
\State Aggregate token features (e.g., via average pooling) to obtain feature vector $F$.
\State Pass $F$ through a SIREN layer: 
\[
HR = \sin\Bigl(\omega_0 \cdot (W F + b)\Bigr)
\]
\State \Return $HR$.
  \end{algorithmic}
\end{algorithm}

Transformer-based architectures, particularly ViT, can effectively analyze satellite imagery and predict weather patterns with high accuracy\cite{garnot2021}. In Super Resolution, these architectures enhance image quality by reconstructing high-resolution images from low-resolution inputs, offering finer details and improved visual fidelity \cite{yang2020l}. It divides the low-resolution image into fixed patches and linearly embeds them into a sequence of tokens. 

\noindent The proposed \textbf{ViSIR} method integrates the strengths of ViT and SIREN to address the SR spectral bias problem: ViT is responsible for learning the long-range dependencies and capturing the global context from the input images \cite{carion2020}, while SIREN captures high-frequency details \cite{SIREN}. These tokens then pass through transformer layers where multi-head self-attention mechanisms and feed-forward neural networks capture global context and long-range dependencies. The final step of the ViSIR pipeline is the SIREN architecture instead of the conventional fully connected neural network. Using sinusoidal activation functions, the SIREN captures high-frequency details to refine and enhance the final image's resolution. This ViSIR modeling pipeline preserves global and local information in the model to reconstruct the Earth System Model while preserving high-frequency details and complex patterns. Figure~\ref{fig-Flowchart} illustrates the flow chart of the proposed method from a low-resolution image to a high-resolution one.

\section{Proof Of Concept}
\label{sec-POC}

\PARstart{F}{irst}, we define three measures of algorithmic performance. Then, we compare the original high-resolution image $I_O$ with the image reconstructed $I_R$ to show the proposed algorithm's performance. 

\noindent \textbf{The Mean Squared Error (MSE)} is defined as the mean difference in pixel intensity between $I_O $ image and $I_R $ image in Eq.~\ref{eq-mse}. The $M $ and $N $ are the dimensions of the images (height and width), and $ I_O(i,j) $ and $ I_R(i,j) $ are the pixel values at position $(i,j) $ in the original and reconstructed images, respectively. While we have RGB images and the three channels, the final values are the mean values calculated for all pixels in all channels.
\begin{equation}
\text{MSE} = \frac{1}{MN} \sum_{i=1}^{M} \sum_{j=1}^{N} \left( I_O(i,j) - I_R(i,j) \right)^2
\label{eq-mse} \end{equation}
A higher MSE means a higher mean discrepancy between the original image $I_O$ and the reconstructed image $I_R$. 

\noindent \textbf{The peak signal-to-noise ratio (PSNR)} measures the quality of reconstructed images in image compression and SR tasks. The PSNR is the ratio between the maximum possible pixel intensity of the image and MSE (Eq.~\ref{eq-mse}) for that image in Eq.~\ref{eq-PSNR}. \begin{equation}
\text{PSNR} = 10 \cdot \log_{10} \left(\frac{\text{MAX}^2}{\text{MSE}}\right) \label{eq-PSNR}
\end{equation} A higher PSNR value means better image quality and less distortion or error in the reconstructed image.

\noindent \textbf{The Structural Similarity Index Measure (SSIM)} considers changes in structural information, luminance, and contrast for comparing the original image $I_O $ with the reconstructed on $I_R $ images. The $\mu_{I_O} $ and $\sigma_{I_O}^2 $ are the mean and the variance of the original image $I_O$, and the $\mu_{I_R} $ and $\sigma_{I_R}^2 $ are the mean and the variance of the reconstructed image $I_R$, while $\sigma_{I_O*I_R} $ is the covariance between original image ${I_O}$ and reconstructed image ${I_R}$ in Eq~\ref{eq-SSIM}.

\begin{equation}
\text{SSIM}(x, y) = \frac{(2*\mu_{I_O} \mu_{I_R} + C_1)(2*\sigma_{I_O*I_R} + C_2)}{(\mu_{I_O}^2 + \mu_{I_R}^2 + C_1)(\sigma_{I_O}^2 + \sigma_{I_R}^2 + C_2)} \label{eq-SSIM}
\end{equation} $ C_1 $ and $ C_2 $ are constants to stabilize the division in Eq~\ref{eq-SSIM}. The SSIM value ranges from -1 to 1, where one means two images are identical, zero means two images have no structural similarity, and negative values indicate that two images are structurally dissimilar.

\begin{table}[htbp]
\centering
\caption{\textbf{(Max, {\em Mean}, Min) values of MSE, PSNR (dB) and SSIM for original $I_O$ and reconstructed $I_R$ images for three measurements and four models.}}
\label{tab-compare}                   
\setlength{\tabcolsep}{3pt}
\begin{tabular}{lccc}                
  \textbf{Measurement $\rightarrow$}
    & \textbf{MSE}
    & \textbf{PSNR (dB)}
    & \textbf{SSIM [0,1]} \\
  \midrule
  \textbf{Model $\downarrow$} & \multicolumn{3}{c}{\bfseries Source Temperature} \\  
  \midrule
  & Max | Mean | Min& Max | Mean | Min& Max | Mean | Min\\
  \midrule
  ViT    
        & 1.54, {\em0.49}, 0.42  
        & 24.32, {\em23.27}, 18.23  
        & 0.66, {\em0.54}, 0.48 \\
  SIREN  & 2.02, {\em0.93}, 0.56  & 21.54, {\em20.21}, 17.36  & 0.61, {\em0.57}, 0.40 \\
  SRGANs & 1.51, {\em0.61}, 0.49  & 22.92, {\em21.43}, 18.00  & 0.64, {\em0.60}, 0.48 \\
  SRCNN  & 0.53, {\em0.14}, 0.05  & 31.87, {\em27.42}, 22.72  & 0.85, {\em0.74}, 0.64 \\
  \textbf{ViSIR}
         & \textbf{0.42, {\em0.13}, 0.06}
         & \textbf{32.10, {\em28.50}, 23.90}
         & \textbf{0.85, {\em0.76}, 0.62} \\
  \midrule
   & \multicolumn{3}{c}{\bfseries Shortwave heat flux}\\ 
  \midrule
ViT &1.18 , {\em 0.42} , 0.31 & 24.55 , {\em 23.92} , 18.74 & 0.69 , {\em 0.66} , 0.52 \\
SIREN &1.43 , {\em 0.73} , 0.62&21.52 , {\em 20.57} , 17.93& 0.62 , {\em 0.58} , 0.49 \\
SRGANS & 1.54 , {\em 0.67} , 0.56&22.16 , {\em 21.22} , 18.16&0.62 , {\em 0.61} , 0.49 \\
SRCNN &0.54 , {\em 0.20} , 0.06&32.34 , {\em 26.89} ,  22.51 & 0.87 , {\em 0.72} , 0.63 \\
\textbf{ViSIR} &\textbf{0.38 , {\em 0.14} , 0.04}&\textbf{33.12 , {\em 28.01} , 24.56}&\textbf{0.87 , {\em 0.75} , 0.66} \\
\midrule
& \multicolumn{3}{c}{\bfseries Longwave heat flux}\\
\midrule
ViT & 1.38 , {\em 0.73} , 0.42&23.87 , {\em 20.55} , 18.28&0.66 , {\em 0.58} , 0.49\\
SIREN & 1.55 , {\em 0.67} , 0.56&22.03 , {\em 21.20} , 18.14&0.62 , {\em 0.60} , 0.49 \\
SRGANS & 1.52 , {\em 0.73} , 0.56&22.10 , {\em 20.57} , 17.91&0.64 , {\em 0.58} , 0.48 \\
SRCNN &0.52 , {\em 0.22} , 0.06& 31.64 , {\em 26.21} ,  21.83 & 0.84 , {\em 0.71} , 0.60 \\
\textbf{ViSIR} &\textbf{0.41 , {\em 0.08} , 0.04}&\textbf{33.56 , {\em 30.50} , 24.24}&\textbf{0.88 , {\em 0.81} , 0.66} \\
  \bottomrule
\end{tabular}
\end{table}

\begin{figure}[!ht] 
 \centering
 \includegraphics[width=\linewidth]{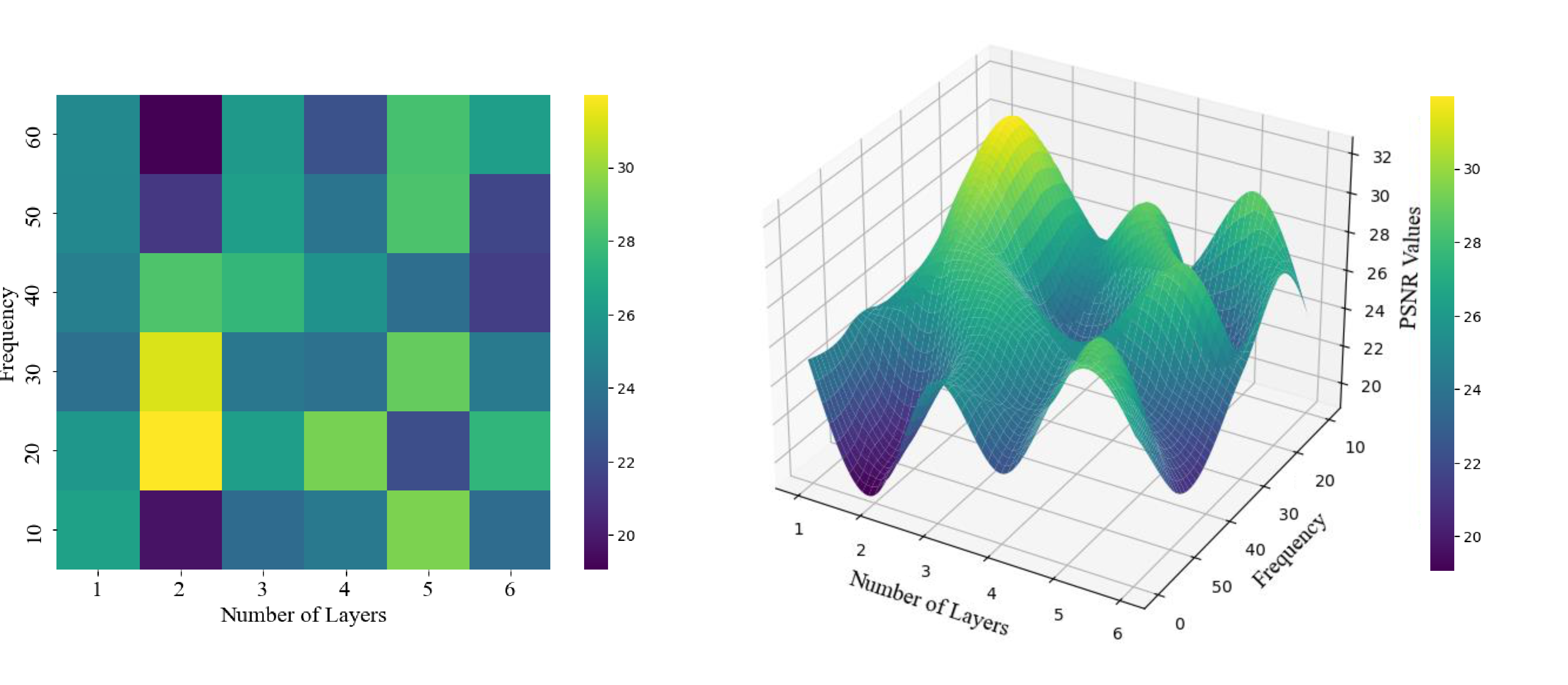} 
 \caption{2D (left) and 3D (right) illustration of the PSNR values for different Frequencies and different numbers of hidden layers used in the proposed ViSIR applied to 180 images of the Surface Temperature variable. }
 \label{fig-FreqVsLayer} 
\end{figure} 

\section{Benchmark Dataset}
\label{sec-data}

In this research, we evaluate our findings using the image set extracted from the Energy Exascale Earth System Model (E3SM) simulation outputs. The E3SM model is an open-access, state-of-the-art, fully coupled model of the Earth's climate, including critical bio-geochemical and cryospheric processes \cite{E3SM2018}. This interpolated model data E3SM-FR is mapped from the E3SM original non-orthogonal cubed-sphere grid simulation data to a regular $0.25^{\circ} \times 0.25^{\circ}$ longitude-latitude grid using a bilinear method. Next, the E3SM-FR data is interpolated onto a $1^{\circ} \times 1^{\circ}$ grid using a bicubic (BC) interpolation method \cite{Passarella2022}. 

\begin{figure}[!ht]
\centering
\begin{minipage}{0.49\linewidth} 
 \includegraphics[width=\linewidth]{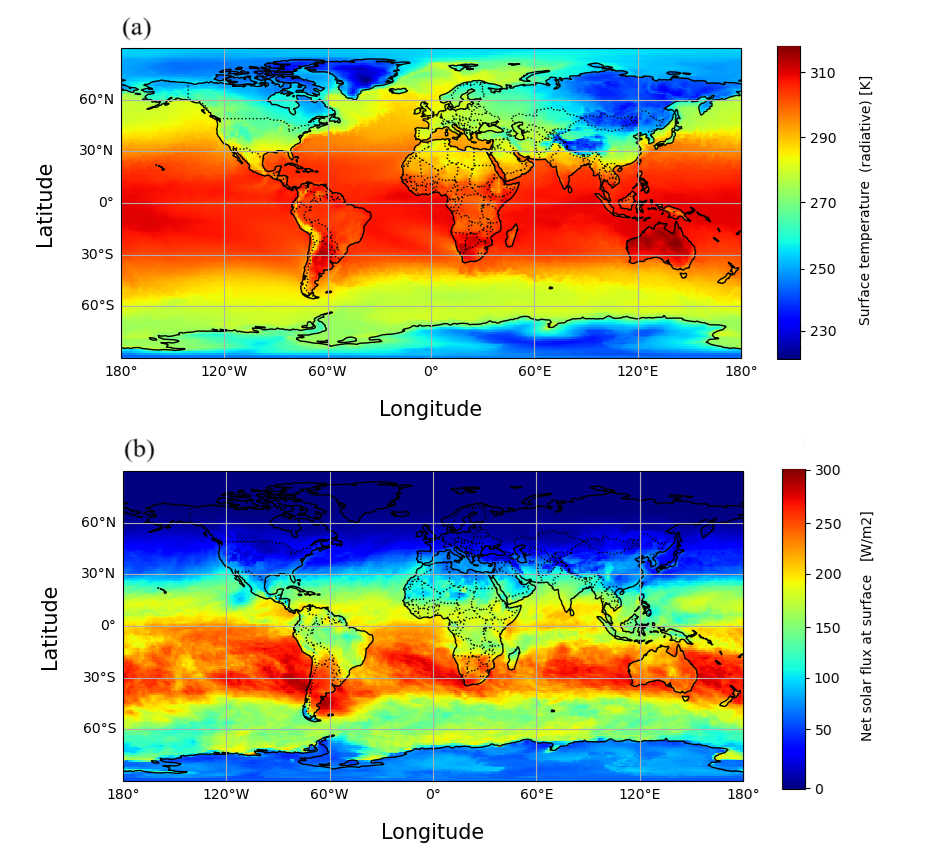} 
\end{minipage}
\hfill 
\begin{minipage}[t]{0.49\linewidth} 
 \raggedleft
 \includegraphics[width=\linewidth]{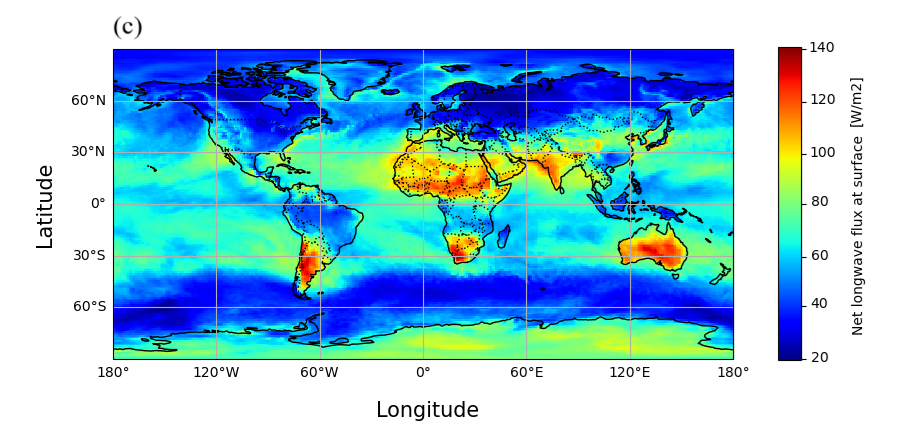}
 \caption{Panels (a), (b), and (c) show surface temperature, shortwave heat flux, and longwave heat flux, respectively, for the first month of year one obtained from the global fine-resolution configuration of E3SM\cite{Nikhil2024}.}
 \label{fig-Motivation}
\end{minipage}
\end{figure}

In our image dataset, each grid point is a pixel, and the entire grid is a high-resolution image with dimensions of $720 \times 1440$ pixels. We derive the R G and B components from the normalized values of surface temperature, shortwave, and longwave heat fluxes data at specific grid points, respectively, as illustrated in Figure~\ref{fig-Motivation}. Note that the corresponding coarse-resolution data has dimensions of $180 \times 360$ pixels, simulating the 4x upsampling. Next, we split the big image into 18 non-overlapping images. The fine-resolution images are now $240 \times 240$ pixels in size, and coarse-resolution images are now $60 \times 60$ pixels in size\cite{Nikhil2024}. In our current dataset, we have $10$ months $\times 18$ images per one of the three measures, a total of 540 images. 

\section{Experimental Results}
\label{sec-Exp}

\PARstart{I}{n} this section, we conduct experiments to compare ViSIR performance with that of ViT \cite{Dosovitskiy2020}, SIREN \cite{SIREN}, and SRGANs \cite{ledig2017}. We use the University's LEAP2 (Learning, Exploration, Analysis, and Process) Cluster processing for the model training and evaluation. The LEAP2 Dell PowerEdge C6520 Cluster enjoys 108 compute nodes, each with 48 CPU cores via two (24-core) 2.4 GHz 6336Y Intel Xeon Gold (IceLake) processors. With 256 GBs of memory and 400 GBs of SSD storage per node, the compute nodes provide an aggregate of 27 TBs of memory and 42 TBs of local storage. We have used 48 CPU cores, with 256GB RAM and 800GB SSD, to run the methods. We applied the hyper-parameter searches with different sinusoidal activation function frequencies ranging from 10 to 60 Hz while varying the number of hidden layers in the SIREN from 1 to 6 to get the best hyper-parameters. Figure~\ref{fig-FreqVsLayer} illustrates the effect of the changing hyper-parameters and the best parameters based on the mean PSNR values. The best PSNR value for EMS images is two hidden layers with a frequency of 20. These are the hyper-parameters used for the rest of the methods in frequency and layers to perform a fair comparison.

\subsection{Experiment 1: Performance Comparison} 
\begin{figure}[!ht]
 \centering
 \includegraphics[width=0.48\textwidth]{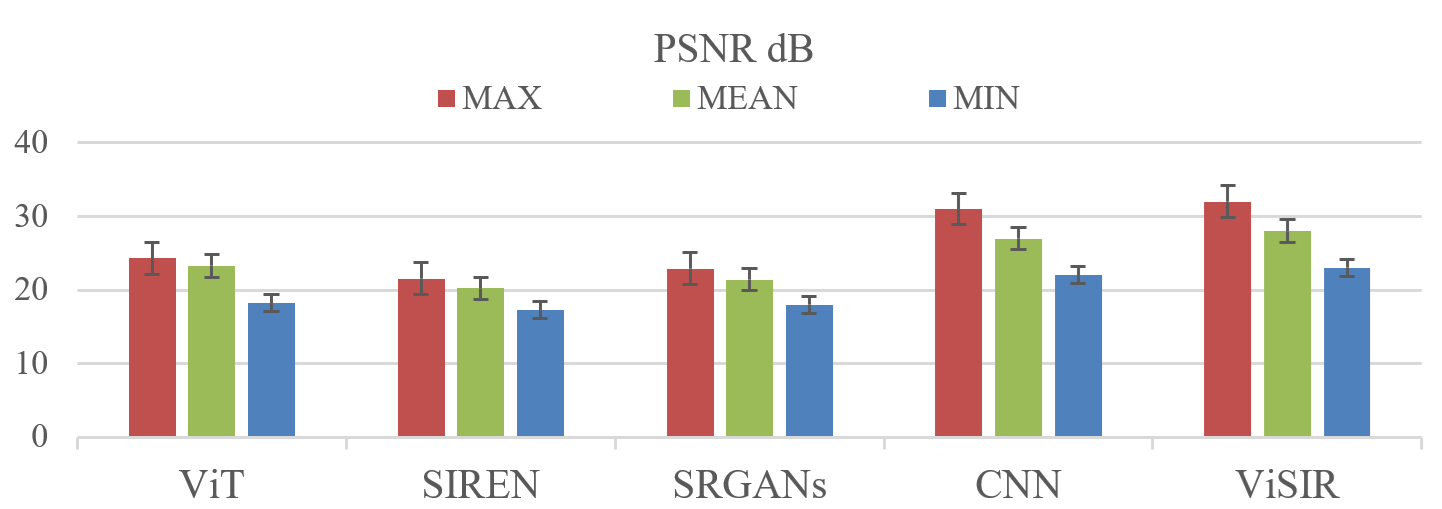}\\ 
 \includegraphics[width=0.48\textwidth]{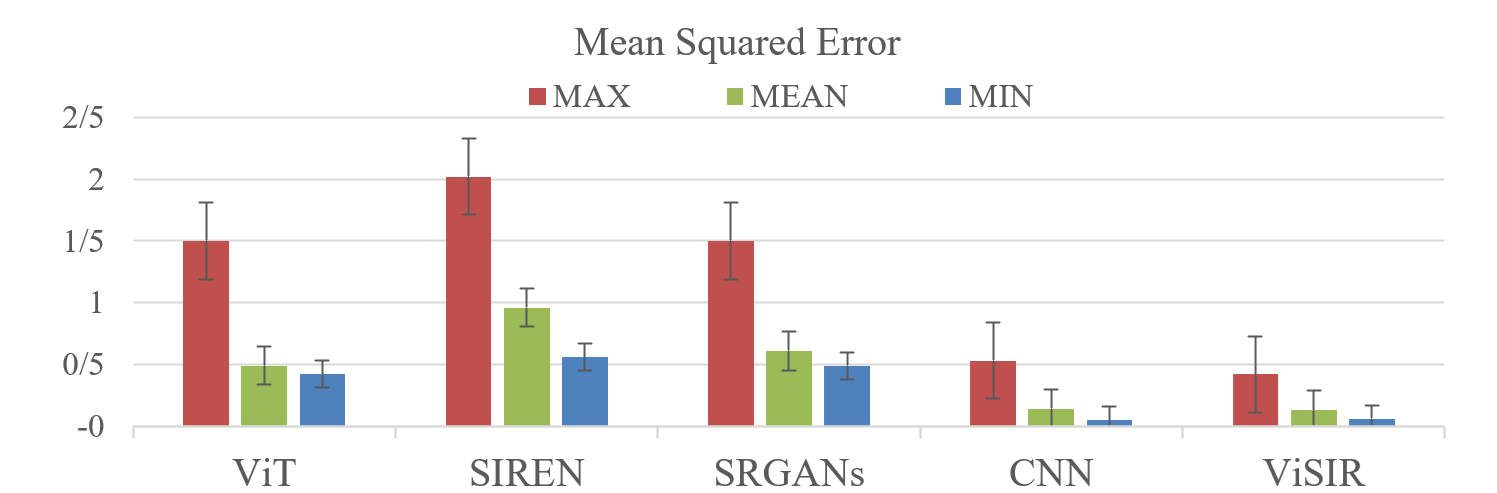}\\
 \includegraphics[width=0.48\textwidth]{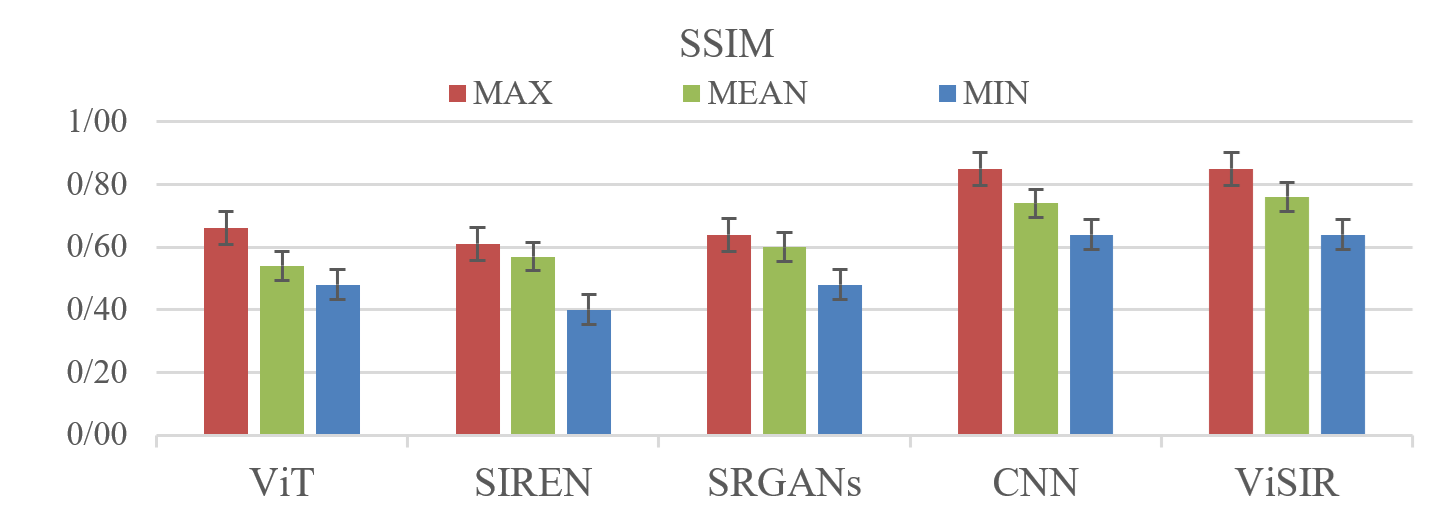}
  \caption{Max Mean, and Min PSNR, MSE, and SSIM values over Source Temperature measurements.}
 \label{fig-Mean}
\end{figure}

This experiment compares ViSIR to ViT, SIREN, CNN, and SRGANS models regarding PSNR, MSE, and SSIM (Figure~\ref{fig-Mean}) scores for three measurements. Table~\ref{tab-compare} summarizes the superiority of the ViSIR performance against three state-of-the-art (SOTA) techniques, ViT, SIREN, and the state-of-the-art SRGANS. ViSIR exhibits the highest PSNR and SSIM for all three measurements, as illustrated in Figure~\ref{fig-Mean}. 

First, the ViSIR improvement over SOTA SIREN is over 10.6dB PSNR and 35.9\% in SSIM improvement. Second, if we compare ViSIR to the next best performer, ViT, it results in 7.8 dB in PSNR improvement and 28\% SSIM improvement. In summary, the strength of ViSIR compared to SOTA for the task is that we combine transformers with the SIREN structure to address the spectral bias that GANS failed to tackle for the single-resolution image reconstruction task. These results highlight ViSIR's superiority in capturing the higher-frequency components in the images and reconstructing high-resolution output effectively.

\subsection{Experiment 2: Corner Case Reconstruction}

In this experiment, we evaluate the corner case scenarios to assess the ViSIR model's performance in handling challenging scenarios. Figure~\ref{fig-Example} (left) is the image in the Surface Temperature dataset with the highest ViSIR PSNR of 32.1 dB. We see that ViSIR handles high-frequency spectral bias well, and the ViSIR algorithm can achieve a notably high PSNR for an image featuring pronounced edges and abrupt color transitions. Figure~\ref{fig-Example} (right) is the image in the dataset with the lowest ViSIR PSNR of 23.9 dB. Table~\ref{tab-compare} summarizes the numerical scores of the maximum (Max), average (Mean), and minimum (min) values per method per evaluation measure. ViSIR is superior in all corner cases as its mean MSE associated with all three variables is lower than the best MSEs of ViT, SRGANs, and SIREN, and its best MSE of 0.08\% is lower than the next MSE score by more than 100\% (0.42\% MSE). SIREN and SRGANS paint almost the same picture regarding MSE, PSNR, and SSIM, while the SRGANs illustrate better results. The ViSIR best PSNR is \textbf{36.7\%} better than ViT.

\begin{figure*}[!ht]
 \centering
 \includegraphics[width=\textwidth]{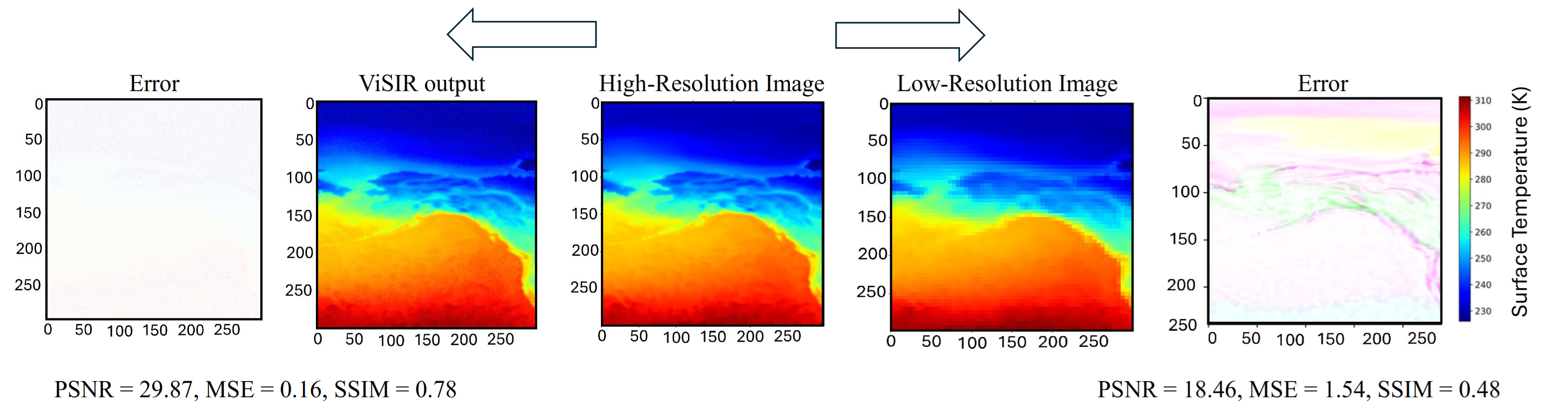} 
 \caption{Image reconstruction from low-resolution Surface Temperature image using ViSIR.}
 \label{fig-Visual}
\end{figure*}
\subsection{Experiment 3: Reconstruction}

\begin{figure}[!ht] 
 \centering
  \centering
  \includegraphics[width=\linewidth]{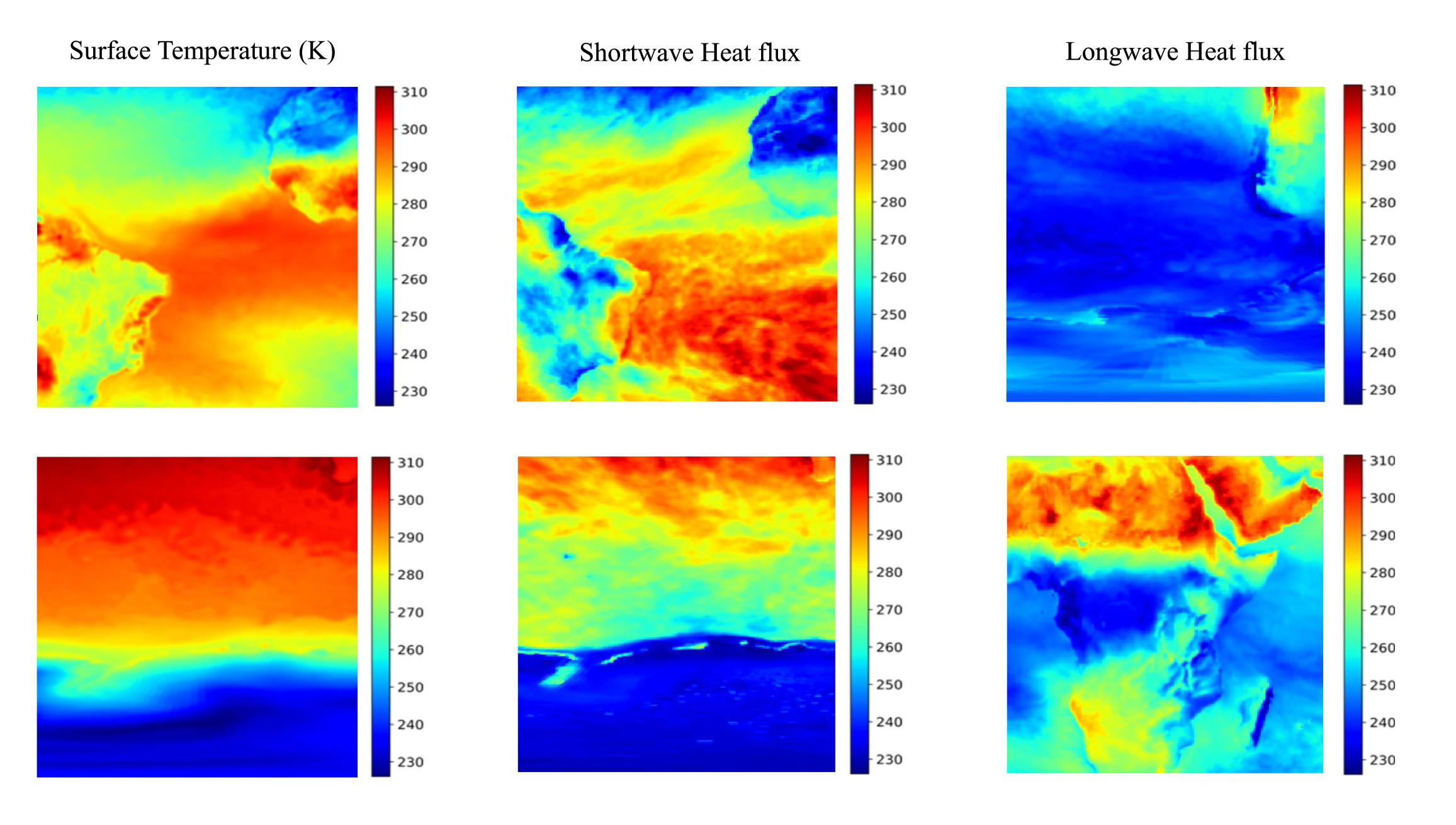}
 \caption{The input images with the highest (Up) and the lowest (Down) PSNR values associated with each variable.}
 \label{fig-Example}
\end{figure}

The single-image SR Task for Reconstruction decomposes the large, high-resolution image into the low-resolution image and the model. Figure~\ref{fig-Visual} illustrates the potential application of the proposed ViSIR model by comparing the low-resolution and ViSIR output in terms of MSE, PSNR, and SSIM. This illustration visualizes the performance of the ViSIR in capturing high-frequency components of the image at the edges, where sharp changes in image intensity are observed. It ensures the capability of the efficient replacement of ViSIR and low-resolution images with 4X high-resolution photos. Table \ref{tab-compare} summarizes the model performance in terms of best, mean, and worst results for all three measures. The findings indicate that the algorithm is effective when processing images containing relatively high-frequency components. Figure~\ref{fig-Example} illustrates input images associated with the best and worst PSNR values achieved by ViSIR.

In summary, the PSNR and SSIM value range comparison in Figure~\ref{fig-Mean} illustrates the dominance of ViSIR for a single image reconstruction task. Additionally, Figure~\ref{fig-Visual} demonstrates the performance of the ViSIR in reconstructing the Surface Temperature high-resolution image from a low-resolution one by addressing the spectral bias issue associated with Neural Network-based algorithms. The error on the right side of this figure indicates the capability of the proposed algorithm in reconstructing high-frequency components of the image found at the edges, while most of the edges are reconstructed by the ViSIR.

\section{Discussion}
\label{sec-discussion}

\PARstart{T}{he} ViSIR framework demonstrates a promising advance in the single-image SR for ESM. By combining the global context modeling capability of ViT with the high-frequency detail preservation offered by SIREN, the proposed method effectively addresses the spectral bias common in traditional SR approaches. Experimental results indicate that ViSIR significantly outperforms standalone ViT, SIREN, SRCNN and SR-GAN models, achieving improvements of up to 9.93 dB in PSNR and substantial gains in SSIM and MSE. Furthermore, the results show an improvement of up to 4.29 dB compared to SRCNN as the second-best performer. These enhancements suggest that our hybrid approach is better at recovering fine details critical for accurate climate modeling and analysis.

One of ViSIR's primary strengths is its ability to leverage the self-attention mechanism inherent in transformers. This allows the model to capture long-range dependencies within the low-resolution input images, which is vital given the complex spatial interactions present in ESM data. The subsequent integration of SIREN layers ensures that the high-frequency components, often lost during down-sampling, are effectively reconstructed, resulting in a more faithful high-resolution output.

However, our study also revealed areas for improvement. One challenge lies in computational efficiency. Although ViSIR achieves impressive reconstruction quality, integrating transformer blocks with SIREN layers increases model complexity and computational cost. Future work should optimize the architecture to balance performance with efficiency, possibly by exploring parameter reduction strategies or employing efficient transformer variants.

Moreover, while the current work centers on single-image SR reconstruction, extending the framework to handle multi-image inputs or video sequences could enhance the model’s utility, especially in dynamic climate prediction scenarios. Future studies will also explore integrating uncertainty quantification methods to assess the reliability of the higher reconstructed outputs.

\section{Conclusion and Future Work}
\label{sec-conclusion}

\PARstart{I}{n}In this paper, we introduce ViSIR, a new approach for high-resolution image reconstruction using Earth System Models (ESM). ViSIR's superior performance lies in combining ViT's global context modeling with SIREN's high-frequency representation capabilities. The proof of concept comparison on images constructed from ESM simulations shows that ViSIR outperforms three state-of-the-art methods significantly in low MSE and higher PSNR and SSIM measures. The ViSIR approach mitigates the spectral bias challenge and produces negligible reconstruction error. Future work will extend ViSIR to multiple images of SR reconstruction and reduce the model's footprint for practical scenarios.

\bibliographystyle{IEEEtran}
\bibliography{Bib/INR,Bib/TesicPub}

\EOD

\end{document}